\def\year{2021}\relax
\documentclass[letterpaper]{article} % DO NOT CHANGE THIS
\usepackage{aaai21}  % DO NOT CHANGE THIS
\usepackage{times}  % DO NOT CHANGE THIS
\usepackage{helvet} % DO NOT CHANGE THIS
\usepackage{courier}  % DO NOT CHANGE THIS
\usepackage[hyphens]{url}  % DO NOT CHANGE THIS
\usepackage{graphicx} % DO NOT CHANGE THIS
\usepackage{natbib}  % DO NOT CHANGE THIS AND DO NOT ADD ANY OPTIONS TO IT
\usepackage{caption} % DO NOT CHANGE THIS AND DO NOT ADD ANY OPTIONS TO IT
\usepackage{amsfonts}
\usepackage[switch]{lineno}  %

\usepackage{xcolor}
 \usepackage{ulem}
 \usepackage{flushend}

\usepackage{newtxtext,newtxmath}
\title{Exploring Explainable  Selection to Control Abstractive Summarization}
 
\author{

    Wang Haonan \textsuperscript{\rm 1}, 
    Gao Yang \textsuperscript{\rm 1}\thanks{Corresponding Author.},
   Bai Yu \textsuperscript{\rm 1},
    Mirella Lapata\textsuperscript{\rm 2},
   Huang Heyan \textsuperscript{\rm 1}
    \\
}

\affiliations{
    \textsuperscript{\rm 1}School of Computer Science and Technology, Beijing Institute of Technology\\
 \textsuperscript{\rm 2}Institute for Language, Cognition and Computation, 
School of Informatics, University of Edinburgh\\
\{hnwang, gyang, yubai\}@bit.edu.cn, mlap@inf.ed.ac.uk, hhy63@bit.edu.cn
}
 
\begin{document}
%\linenumbers  %
\maketitle

\begin{abstract}
 Like humans, document summarization models can interpret a document’s contents in a number of ways. Unfortunately, the  neural models of today are largely black boxes that provide little explanation of how or why they generated a summary in the way they did. Therefore, to begin prying open the black box and to inject a level of control into the substance of the final summary, we developed a novel select-and-generate framework that focuses on explainability. By revealing the latent centrality and interactions between sentences, along with scores for sentence novelty and relevance, users are given a window into the choices a model is making and an opportunity to guide those choices in a more desirable direction. A novel pair-wise matrix captures the sentence interactions, centrality and attribute scores, and a mask with tunable attribute thresholds allows the user to control which sentences are likely to be included in the extraction. A sentence-deployed attention mechanism in the abstractor ensures the final summary emphasizes the desired content. Additionally, the encoder is adaptable, supporting both Transformer- and BERT-based configurations. In a series of experiments assessed with ROUGE metrics and two human evaluations, ESCA outperformed eight state-of-the-art models on the CNN/DailyMail and NYT50 benchmark datasets. 
\end{abstract} 
\section{Introduction}
\label{sec:intro}
The ability to generate summaries of documents is a valuable tool over the past several years, and  neural networks have been responsible for a step-change in the quality of  both extractive and abstractive summarization. Extractive methods simply draw out and concatenate the key topic sentences in a document \citep{nallapati2017summarunner,zheng2019sentence},  while abstractive techniques reorder words and sentences and even generate new language to, hopefully, produce a concise and eloquent piece of the given content  \citep{DBLP:conf/acl/SeeLM17,celikyilmaz-etal-2018-deep,wenbo2019concept}. % Despite their success, it remains a challenging task to model long-range context for document summarization.
However, despite recent advancements, modelling concepts that span more than a few sentences, i.e., long-range contexts, still a challenging task. Moreover, current models provide little to no explanation of the interpretation they took away from parsing a document and why they chose to summarize its content in the way that they did.

Currently, two broad strategies for tackling this problem are explored. The first is to use pre-trained language model, such as ELMo \citep{peters2018deep}, OpenAI GPT \citep{radford2018improving} and BERT \citep{devlin2018bert}, have achieved state-of-the-art performance on long-range  contextual learning   and various NLP tasks, such as QA \citep{xu-etal-2019-bert} and summarization  \citep{liu2019text,zhang-etal-2019-hibert}. 
%There is an obstacle that it only accepts a maximum length of 512 tokens. As a result, to digest long-range of the document, some tricks are often adopted, such as truncating to the predefined tokens of the document  \citep{RenHWLBWG19,dong2019unified}, adding  position embeddings that are %initialized randomly and 
%fine-tuned in the encoder \citep{liu2019text}. 
The other idea is to use  a  \textit{select and generate} framework, where an extractor selects salient sentences, then an abstractor  generates a summary. The most recent frameworks based on this hybrid paradigm either  follow a  two-stage pipeline~\citep{chen2018fast,sharma2019entity} or an end-to-end learning approach~\citep{hsu2018unified,shen2019select,gehrmann2018bottom}. 
The most appealing advantage is  to explicitly obtain desirable content of the sources, such as entity-aware selection \citep{sharma2019entity} or word selection through latent switch variables \citep{gehrmann2018bottom,shen2019select}.

These approaches perform abstractive summaries which largely rely on selecting informative content  (extractor) as well as aggregating into a summary  in line with linguistic expression (abstractor). But, currently, the extractors are largely black-box decisions without a rationale of what is informative content.   \citet{peyrard2019simple}  proposed rigorous definitions of the concepts in summarization, including \textit{redundancy}, \textit{relevance} and \textit{informativeness}. While, more in-depth investigation of these concepts are needed for them to be truly useful to document summarization. For instance, we need to better understand inter-relations between sentences with respect to these attributes. We need methods for  identifying the sentence \textit{informativeness}, identifying whether a sentence is \textit{relevant} to a document and, if so, to what extent. Another importance influence is the \textit{novelty} of the  contribution  a sentence makes to a summary. 

Moreover, abstractive summarization suffers from a  major problem known as   hallucination, where the model generates   fictional content  \citep{maynez2020faithfulness}. The cause is believed to due to misrepresenting content in a batch of input documents and fusing concepts across those documents when generating abstractive summaries. Additionally, some of the new terms introduced are thought to come from background knowledge, not from the current inputs.  Some researchers have attempted to alleviate this problem with pointer mechanisms to desired content  \citep{DBLP:conf/acl/SeeLM17,wenbo2019concept,celikyilmaz-etal-2018-deep} or by interpolating nearest neighbors computed from the inputs \citep{khandelwal2019generalization} and so on. However, due to the scattered tracts of information in long documents, it is inevitable that some irrelevant and unnecessary content will be picked up when generating summaries. As a result, there is potential for an abstractive summary to completely depart from the gold summary into a fictional hallucination and, unfortunately, this is difficult to control.  

Therefore, to reveal more of the inner workings of these black-box models so as to inject a level of control into the substance and integrity of the final summary, we developed a novel select-and-generate framework, called \textbf{ESCA} (Loosely, means Explainable Selection module to Control the generation of Abstractive summaries), that focuses on explainability. The key to the framework is an interaction matrix that highlights the decisions made about each sentence, which can be decoupled into three explicit components, the informativeness of a sentence, its relevance to the substance of the document, and its novelty with respect to the accumulated summary representation. A novel pair-wise ranking extractor then selects sentences for extraction, favoring the complex relations within each sentence pair and its potential influence over the summary. To avoid hallucinations,  a sentence-deployed attention mechanism in the abstractor, but populated with values from the extractor, ensures the abstractive summary focuses on both correct and desired concepts. As such, the extractor and abstractor are seamlessly integrated with a  deployment-based pointer in an end-to-end manner. Further, which content is selected for extraction can be controlled by setting thresholds for novelty and relevance and applying a mask that adjusts the probability of extraction accordingly.   

In summary, our contributions include:  1) an explainable content selection module for document summarization; 2) the ability to extract the appropriate content for generating a desired summary based on explicit and quantified measures of informativeness, novelty and relevance to the final summary;   
 3) automatically creating  synthetic datasets w.r.t novelty and relevance for exercising controllable inference without the need to retrain the entire system. 
A series of experiments assessed with ROUGE metrics and two human evaluations demonstrate that ESCA provides summaries of higher quality than eight state-of-the-art models on the CNN/DailyMail and NYT50 benchmark datasets.

\section{Related Work}
\label{sec:relatedwork}
As opposed to extractive summarization, where all but the most salient and meaningful sentences are removed to reduce an entire document down to a short summary of its contents, abstractive summarization generating new or rephrased words and sentences to produce the summary. 
%supposedly digests and understands the source content and  generates a new order of words and shorter sentences.
Headline generation is a subtask of abstractive summarization and, in this area, seq2seq models have largely accomplished the goal of generating  snappy and expressive headlines. 
%substantially succeeded in achieving it  
~\citep{DBLP:conf/conll/NallapatiZSGX16,DBLP:conf/acl/ZhouYWZ17,shen2019select,wenbo2019concept}. However, summarizing content with notions that span more than a few sentences, i.e., long-range contexts, with abstractive techniques remains a significant challenge. 
%it remains a significant challenge to deal with the long-range context of document summarization in an abstractive manner. 
%The reasons are quite complex. One is derived  from semantic representations which determine  whether the model is able to accurately  understand the document. This has been intensively studied in the community of pre-trained models~\citep{peters2018deep,radford2018improving,devlin2018bert,liu2019text,zhang-etal-2019-hibert}, which is not the focus  of this paper.
%The main reason is that today's language generation models are often a ``black-box" neural networks that offer no explanations for how and why they make choices they do. Obviously, it is difficult to correct problems when we do not fully understand where the models are going wrong.     

%Therefore, selecting desired and meaningful content has been an urgent task. 
Models called pointer-generator overcome this explainability  problem to some extent by using attention as a pointer, conditioned on contextual information, jointly determine which language to select/generate based on probability \citep{vaswani2017attention,DBLP:conf/acl/SeeLM17}. 
Further, pointer generators can operate at either the word level \citep{wenbo2019concept,DBLP:conf/naacl/CelikyilmazBHC18} or the  sentence level~\citep{chen2018fast, sharma2019entity}.  At the word level, 
\citet{DBLP:conf/acl/ZhouYWZ17} used soft gating on the source document to produce summaries, while \citet{gehrmann2018bottom} pre-trained a sequential word selector to constrain attention from the source document. \citet{hsu2018unified} updated word attention by considering importance at the sentence level. Among the sentence-level models, \citet{tan2017abstractive} used a graph-based attention mechanism with an abstractive model leveraged by improving salient sentence selection. 
\citet{li2018improving} achieved the same goal with  an information selection layer consisting of global filtering and sentence selection modules.   
\citet{you2019improving} subsequently improved  salience attention by introducing a Gaussian focal bias to better inform the selection process. 

%Regarding 
In terms of the generation process, a single text can be summarised in diverse target sequences with different focus \citep{cho2019mixture}. To tackle this issue, \citet{shen2019select} used decoupled content selection to allow fine-grained control over the generation process.  
In our framework, we leverage the benefits of a  pointer generator model for selection, but we also explore explaining the content to be selected for extracted as a way to control the generation process so as to produce a desirable summary.

\section{Background: Encoder-Decoder Framework}
\label{sec:background}

%Formally, the abstractor can be described as a  generation process where a sequential input is summarized into a shorter sequential output through a neural network. 
In summarization, consider a sequential input  ${X} = \lbrace  \boldsymbol{x}_1, \dots,  \boldsymbol{x}_j, \dots,   \boldsymbol{x}_n \rbrace$   of  $n$ number of words with  $j$ as the index of the input, the shortened output, i.e., the summary, is denoted as $ {Y} = \lbrace  \boldsymbol{y}_1, \dots,  \boldsymbol{y}_t, \dots, \boldsymbol{y}_m \rbrace$ with $m$ number of  words, where $t$ indicates the position of the output.  
%The basic structure is based on Transformer. It is composed of a stack of $N$ identical layers, and each layer has two sub-layers: 
% \begin{equation*}
%     \boldsymbol{h}_1^l = \textsc{LayerNorm}(\boldsymbol{h}_2^{l-1}+\textsc{MulH}_{att}(\boldsymbol{h}_2^{l-1}))
% \end{equation*}
% \begin{equation}
%     \boldsymbol{h}_2^l=\textsc{LayerNorm}(\boldsymbol{h}_1^l+\textsc{FFN}(\boldsymbol{h}_1^l))
% \end{equation}
% \begin{equation*}
% \begin{split}
% \boldsymbol{h}_1^l =&  \textsc{LayerNorm}(\boldsymbol{h}_2^{l-1}+\textsc{MulH}_{att}(\boldsymbol{h}_2^{l-1}))\\
%     \boldsymbol{h}_2^l=& \textsc{LayerNorm}(\boldsymbol{h}_1^l+\textsc{FFN}(\boldsymbol{h}_1^l))
% \end{split}
% \end{equation*}
%Encoder-decoder frameworks consist  of  an  encoder  and  an attention-equipped decoder. 
The basic structure is based on Transformer unit composed of a stack of $N$ identical layers, and each layer has two sub-layers: the first is a self-attention sub-layer $\boldsymbol{h}_1^l$, and the second is a  feed-forward sub-layer $\boldsymbol{h}_2^l$ with a   depth of $l$. %\textsc{LayerNorm} is the layer normalization, and the multi-head operation is 
Then, a multi-head operation follows the feed-forward sub-layer. 
% \begin{equation*}
%     \textsc{MulH}_{att}(\boldsymbol{h}_2^{l-1}) = \textsc{Concat}(\boldsymbol{H}_1,\cdots, \boldsymbol{H}_h)\boldsymbol{W}_l
%   \end{equation*}
% where $\boldsymbol{H}_i$ is the self-attention operation of layer $l$ at $i$th head. 
%This operation allows to jointly attend to information from different representation positions. $\boldsymbol{h}_0 = \textsc{PosEmb}(\boldsymbol{x})$ and \textsc{PosEmb} is the positional embeddings of the input. The feed-forward sub-layer takes the output of self-attention sub-layer as the input and computes with a position-wise fully connected feed-forward network:
% \begin{equation}
%     \textsc{FFN}(\boldsymbol{h}_1^l) = \text{ReLU}(\boldsymbol{h}_1^l \boldsymbol{W}_1^l+\boldsymbol{b}_1^l)\boldsymbol{W}_2^l +\boldsymbol{b}_2^l
% \end{equation}
%where $\boldsymbol{W}_l, \boldsymbol{W}_1^l, \boldsymbol{W}_2^l$ are all learnable parameters, and $\boldsymbol{b}_1^l, \boldsymbol{b}_2^l$ are biases. ReLU is hidden activation function.
The final output sequence of the encoder is denoted as $\boldsymbol{Z}_e$. %We also employ the pre-trained BERT encoder  \citep{devlin2018bert} and the implementing details are included in Section \ref{sec:exp}. 
%For both the Transformer-based and BERT-based configurations, we employ a similar decoder equipped with attentions.  
The decoder consists of a similar stack of $N$ identical layers. But, in addition to the two sub-layers for each layer, the decoder includes a third sub-layer, that performs multi-head attention over the output of the encoder stack. Following this procedure, an attention value between the decoder position vector $\boldsymbol{s}_t$ and the encoder sequence output $\boldsymbol{Z}_e$ is calculated for  each source position.  
Attention on the source input at the decoder position $t$ is then calculated with the formula $
    \boldsymbol{\alpha}_t = \text{softmax} (\frac{\boldsymbol{Q} \boldsymbol{K}^T}{\sqrt{d_m}})
$, 
where $\boldsymbol{Q}$ is $\boldsymbol{s}_t$ and $\boldsymbol{K}$ is $\boldsymbol{Z}_e$.  %For simplicity, we just use the first head of the $h$ multi-attentions. 
The context vector at the decoding position $t$ is $\boldsymbol{h}_t^*= \boldsymbol{\alpha}_t \boldsymbol{Z}_e$.  
From here, the decoder  generates a summary, called  the target summary,  from a vocabulary distribution $P_{vocab}(w)$ through the following  process:
\begin{equation}
\label{eq:outy}
\begin{split}
P_{vocab}(w) & =P(y_t|\boldsymbol{y}_{<t},\boldsymbol{x}; \theta ) \\
 =& \text{softmax}(\boldsymbol{W}_2(\boldsymbol{W}_1[\boldsymbol{s}_t, \boldsymbol{h}_t^* ]+\boldsymbol{b}_1)+\boldsymbol{b}_2)
\end{split}
\end{equation}
 
 \section{The Proposed Model}
 
\begin{figure}[t]
	\centering
     \includegraphics[width=0.45\textwidth]{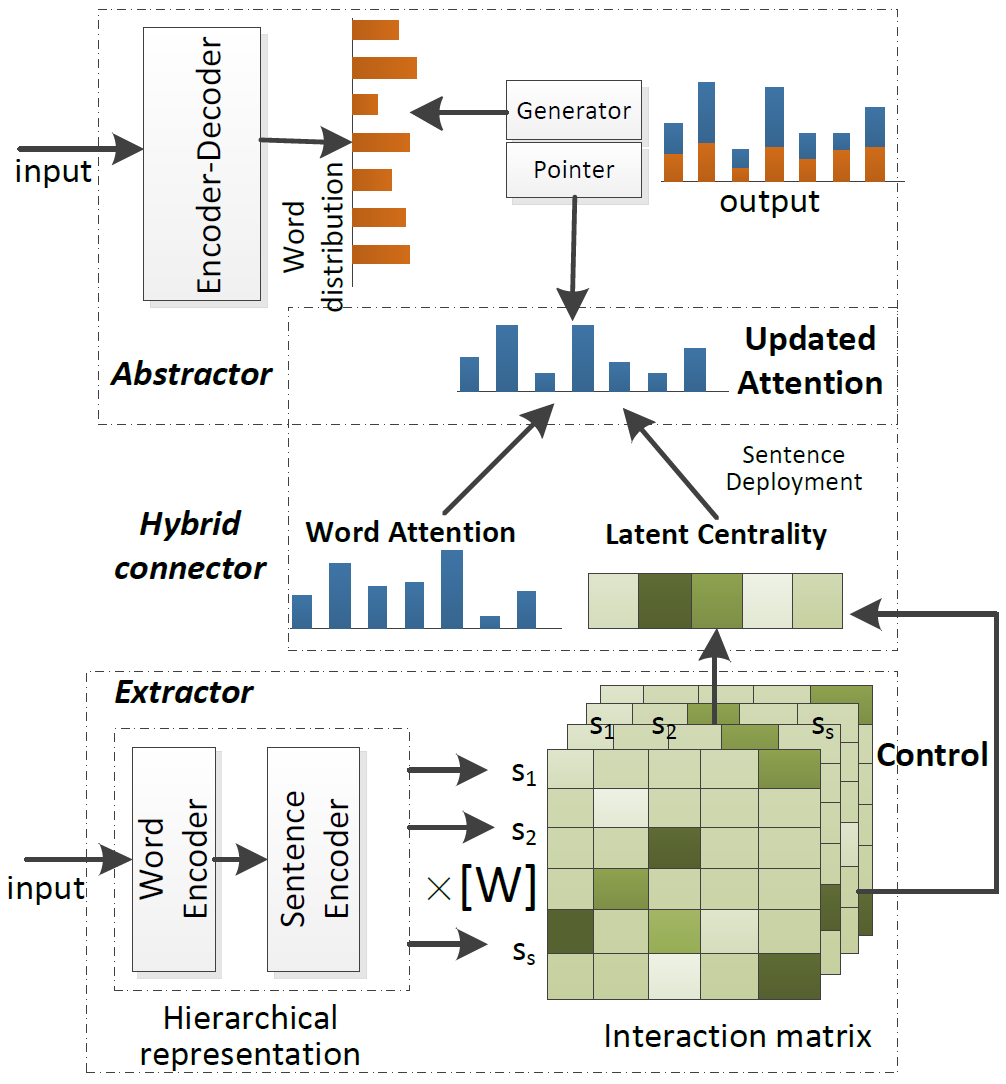}
	\caption{The framework of the proposed method}  
	\label{fig:framework}
\end{figure}

Our model is an end-to-end hybrid summarization framework. The architecture and summarization process is demonstrated in Figure \ref{fig:framework}. It comprises: (1) a pair-wise extractor that incorporates a sentence interaction matrix and uses  latent centrality; (2) the abstract generation is guided by the sentence deployed attention as a pointer to hybrid with a Pointer Generator (PG) abstractor and (3) controllable and tunable interaction matrix that explains selecting different content  that will affect the final abstractive summary.  

\subsection{The Extractor}
\label{sec:modelextact}
The encoder of our framework can be equipped with any vector-based neural networks. In this paper, we implement Transformer and BERT, showing its  flexibility. The detailed settings are depicted in  the experimental section.   %Specifically, 1) the Transformer-based sentence and document are represented by hierarchical Transformer architecture \citep{vaswani2017attention}. 2) Similar to \citep{liu2019text}, several inter-sentence Transformer layers are stacked on top of BERT output.%, then  BERT-based representations are collected upon two stacked Transformer layers.  
 \subsubsection{Explaining the Interaction Matrix}
 \label{sec:matrix}
As stated by  \citet{nallapati2017summarunner}, complex relations exist in each sentence pair that add informativeness, novelty and relevance to the content of a document. Inspired by  the relations, our  sentence interaction matrix,  $\boldsymbol{Q}^{s}$, reflects these complexities along with the similarity of the sentence pair. In  $\boldsymbol{Q}^{s}$, $s$ denotes the number of sentences. We further note that ``interactions" have a  direction, because, as suggested by  \citet{zheng2019sentence}, 
the contribution induced by two sentences' relations to their respective importance as a summary can be unequal. Furthermore, the mutual influence of the sentence pair may have a  different direction,  which is  grounded in the  theory of discourse structure  \citep{mann1988towards}. The directional influence of sentence $j$ to sentence $i$ is denoted as $q_{ij}$ in the interaction matrix  $\boldsymbol{Q}^{s}$, including informativeness of the sentence $s_i$, relevance of $s_i$ to the document and  novelty of $s_i$ to existing  summary. 

Specifically, the \textit{informativeness} refers to how important and informative of a specific sentence $i$, the \textit{relevance} indicates to what extent the sentence $i$ is relevant to a document $d$, and the \textit{novelty} means what the new information of the sentence $i$ contributes to the summary. Therefore, the above attributes explicitly decouple the interaction matrix, showing  explainable meanings to the summary, and the directional influence score $q_{ij}$ in $\boldsymbol{Q}^{s}$ is:

%  \begin{equation}
%      \begin{split}
%      &q_{ij}(\boldsymbol{h}_i,\boldsymbol{s}_i,\boldsymbol{h}_j,\boldsymbol{d} )= \sigma(\underbrace{\boldsymbol{W}_c\boldsymbol{h}_i}_\text{informativeness} 
%      +\underbrace{\boldsymbol{h}_i^\mathrm{T}\boldsymbol{W}_s\boldsymbol{h}_j}_\text{similarity}\\
%      &\underbrace{-\boldsymbol{h}_i^\mathrm{T}\boldsymbol{W}_n\tanh(\textbf{nov}_i)}_\text{novelty}
%      +\underbrace{\boldsymbol{h}_i^\mathrm{T}\boldsymbol{W}_r\boldsymbol{d}}_\text{relevance}
%      +\boldsymbol{b}_\text{matrix}
%      )\\
%      \end{split}
%      \label{equ:interQ}
%  \end{equation}
 \begin{equation}
     \begin{array}{ll}
     & q_{ij}(\boldsymbol{h}_i,\boldsymbol{s}_i,\boldsymbol{h}_j,\boldsymbol{d} )= \sigma(\underbrace{\boldsymbol{W}_c\boldsymbol{h}_i}_\text{informativeness} 
     +\underbrace{\boldsymbol{h}_i^\mathrm{T}\boldsymbol{W}_r\boldsymbol{d}}_\text{relevance} \\
     & +\underbrace{\boldsymbol{h}_i^\mathrm{T}\boldsymbol{W}_s\boldsymbol{h}_j -\boldsymbol{h}_i^\mathrm{T}\boldsymbol{W}_n\tanh(\textbf{a}_i)}_\text{novelty}
     +\boldsymbol{b}_\text{matrix}
     ) \\
     \end{array}
     \label{equ:interQ}
 \end{equation}
 % the transformer hier.., W trainable. 
where $\boldsymbol{h}_i$ is the representation of sentence $i$, and $\boldsymbol{d}$ is the vector of the input document.  $\textbf{a}_i$ is the accumulated summary representation w.r.t the current sentence $i$ and is  $\textbf{a}_i =  \frac{1}{s}\sum_{t=1}^{i-1}\sum_{k=1}^{s}\boldsymbol{h}_{t}\times q_{tk}$, where $q_{tk}$ represents the influence of sentence $k$ to sentence $t$. Note that  the novelty roughly decreases with the latter sentences as normal summary positioned in the front position. 
 $\sigma$ is a sigmoid function, and  $\boldsymbol{W}_c, \boldsymbol{W}_s, \boldsymbol{W}_n, \boldsymbol{W}_r $, $\boldsymbol{b}_\text{matrix}$  are trainable parameters.   
%, where the importance score $c_i$ will be described in the Section \ref{centrality}.  

 \subsubsection{Latent Centrality Calculation} 
The interaction matrix $\boldsymbol{Q}^{s}$ stores the mutual influence of each sentence pair, which  helps to estimate the overall importance of the sentence.  
%each sentence pair relation  that enables to  construct a graph of the document.
There are several summarization models for computing the centrality of a  sentence, including  the graph-based TextRank \citep{mihalcea2004textrank} and LexRank. Additionally, \citet{tan2017abstractive}  drew on a similar idea with a model that determines  sentence salience via graph-based  attention. In  our end-to-end setting, the interaction matrix $\boldsymbol{Q}^{s}$ is directly transformed into the sentences' distribution, which is then  converted into a centrality vector:
\begin{equation}
    \boldsymbol{c} = \boldsymbol{Q}^s \boldsymbol{W}_q
    \label{equ:centrality}
\end{equation}
where  $\boldsymbol{c}$ = $[c_1, \cdots,  c_s]\in \mathcal{R}^s$ is the sentence centrality, and $\boldsymbol{W}_q \in \mathcal{R}^{s\times 1}$. In our experiments, we truncated the sentence   number $s$ in the  documents to a maximum of 50. 

 \subsubsection{Pair-wise Learning Extractor}
The process of extraction can be framed as a classification problem. \citet{nallapati2017summarunner,hsu2018unified, liu2019text} all use  a point-wise ranking approach in which sentences are encoded as hidden representations. Then, a binary classifier  is trained on those representations to predict whether or not they are  suitable for the summary. However, because point-wise learning is not yet  powerful enough to accurately reflect the interactions between sentences, we introduced a new pair-wise loss function supported by inter-sentence labels that helps the extractor decide  the summary classification. More specifically, first, each sentence is labeled as described in Appendix A.  Then, the inter-sentence label for each sentence pair $\hat{P}_{ij}$ is marked with $\{ 0,  1\}$, where 1 indicates the sentence $i$  has been selected for the summary, but sentence $j$  has not; 0 indicates the opposite – that sentence $j$ has been selected for the summary while $i$ has not. To adapt our supervised system to summarization, the predicted co-occurrence probability $r_{ij}$ of sentence $i$ needs to be calculated. The formula is  $\sigma(c_i-c_j)$, and the loss function is then defined as
\begin{equation}
    \mathcal{L}_{\text{ext}}= -\sum_{i=1}^m(\hat{P}_{ij} \log {r}_{ij}+(1-\hat{P}_{ij})  \log (1-{r}_{ij}))
\end{equation}
%where let the inter-sentence label $\hat{P}_{ij} \in \{ 0, 0.5, 1\}$.% be defined as 1 if $i$ is preferred over item $j$ as a summary, 0 if  $i$ is less preferred, and 0.5 if they are given the same preference as summary or not.% as the summaries while -0.5 if none of them belongs to the summary. 

\subsection{The Abstractor}
\label{sec:abstract}
The abstractor is based on a pointer-generator network containing two sub-modules: the pointer network and the generation network. These two sub-modules jointly determine the probability that a word will be included in the final generated summary.  
Our proposed model  essentially leverages this configuration that integrates a new sentence deployed pointer, introducing the selected content flow into the generation network in the hybrid framework. 

%Therefore, in our proposed framework, the abstractor is achieved by a pointer-generator  network, which is encoded by a stacked-layer of self-attention mechanism \citep{vaswani2017attention} (i.e., Transformer) and also adaptable with pre-trained BERT encoder  \citep{devlin2018bert,liu2019text}. 
%Our proposed model is essentially an upgrade to this configuration that integrates a new concept pointer network within a unified framework.
\subsubsection{Sentence Deployed Pointer Generator} 

The pointer network uses attention as a pointer to select segments of the input as outputs \cite{vinyals2015pointer}. 
As such, a pointer network is a suitable mechanism for extracting salient information, while remaining flexible enough to interface with a language model for generating an abstractive summary   \cite{DBLP:conf/acl/SeeLM17}. 

In our pointer network, the selected segments of input can be  updated by the extractor with respect to their extractive-oriented centrality of each sentence. 
To influence the sequence generation, sentence importance needs to be deployed to the word  level. The deployment should determine how much information flow is delivered to the word-level generation, at the same time considering the importance of the derived sentence. 
With these values,  the pointer can seamlessly link the extractor with the abstractor via the hybrid connector.

%The generation probability $p_{gen}$  for the generation network \cite{DBLP:conf/acl/SeeLM17} is conditionally learned by contexts of encoder and decoder. The pointer is taken by the attention distribution that will be updated by our proposed  hybrid connector. %The output distribution is finally combined with the pointer of attention distribution and generation of vocabularies.  

%\subsection{Sentence Deployed Attention}
%We first train a PG model on the full dataset as well as the pair-wise sentence extractor defined above.
%\textbf{Hybrid Connector} 
The pointer is taken by the attention distribution that will be updated by our proposed  \textit{hybrid connector}. The  hybrid is achieved by the sentence deployment attention mechanism,  which  controls the generation  process by  focusing on what the selected content  explicitly conveys. The equation for calculating the pointer distribution leveraged by sentence deployed attention is as follows,  
\begin{equation}
\begin{split}
    \hat{\alpha_t^n} &= \frac{\alpha_t^n(1+p_{sen} c_{m_n})}{\sum \alpha_t^n(1+p_{sen} c_{m_n})}\\
     p_{sen}&=\sigma(\boldsymbol{W}_{sel}\boldsymbol{E}^t_{sel}+\boldsymbol{b}_{sen})
\end{split}
\label{eq:focus}
\end{equation}
where $c_{m_n}$ denotes the score of the  sentence $m$ that the word $n$ belongs to.  $\boldsymbol{E}^t_{sel}$ is the representation of the selected sentence $m$ at the decoding step $t$.  
$p_{sen}$ decides the degree of influence a sentence will have on the summary.  $\boldsymbol{W}_{sel}$ is a trainable parameter. 
Additionally, the generation probability $p_{gen}$ is   modified with  
\begin{equation}
   p_{gen} = \sigma(\boldsymbol{W}_{h^*}\boldsymbol{h}^*_t+\boldsymbol{W}_s \boldsymbol{s}_t +\boldsymbol{b}_{gen}) 
\end{equation}
The pointer is taken based on the updated  attention distribution $\hat{\alpha_{t}}$ over the source text, and the final output distribution is combined as follows: 
\begin{equation}
%\begin{split}
    P_{\text{final}}(w) =  p_{gen}P_{vocab}(w)+ (1-p_{gen})(\sum\nolimits_{j:w_j=w}\hat{\alpha_{t,j}}) 
%\end{split}
\end{equation}

The basic generator objective is derived by maximizing the likelihood of  pointer-generator during training, given a reference summary $y^* = \{y^*_1, y^*_2,\cdots, y^*_{m'}\}$ for a document $x$, and the training objective is to minimize the negative log-likelihood of the target word sequence:
\begin{equation*}
%\resizebox{\hsize}{!}{
   \mathcal{L}_{abs} = -\sum_{t = 1}^ {m'} \log P_\text{final}(y_t^*| y_1^*,\cdots, y_{t-1}^*, \boldsymbol{x}) 
    \label{equ:mle}
 %   }
\end{equation*}
Overall, the learning objective is  $\mathcal{L} = \mathcal{L}_{ext} + \mathcal{L}_\text{abs}$.

\subsection{Controllable Inference}
\label{sec:control}
Since the interaction matrices capture the inter-sentence relations leveraged by different explainable aspects, as outlined in the section of extractor, the overall centrality has the potential to reflect the aspects  of the content and, in turn, explain how selecting one sentence over another will influence the final abstractive summary.  Hence, to explore this explainability and fine-tune which sentences to extract to produce the most desirable abstractive summary, the sentence selections can be manipulated through several mask matrices $\boldsymbol{M}$ based on controllable thresholds for novelty $\epsilon_n$ and relevance $\epsilon_r$ versus each sentence’s scores in these areas.  Note that the informativeness is not proper to be controlled since it only relates to the sentence itself without any interaction with other sentences or the document. 
The fine-tuning control is applied using the following equation:    
\begin{equation}
   \hat{\boldsymbol{Q}^s} = \boldsymbol{Q}^s \odot \boldsymbol{M}, \text{ where  } M_{ij} = \left\{\begin{array}{l}1, val \ge \epsilon \\0, val<  \epsilon \end{array}\right.
   \label{equ:control}
\end{equation}
where $\odot$ is element-wise multiplication, and $val$ is the  $\sigma({novelty})$ or $\sigma({relevance})$ calculated from Eq.(\ref{equ:interQ}). In this way, the mask matrices, $\boldsymbol{M}_n$ (novelty) or $\boldsymbol{M}_r$ (relevance), can be adjusted to control which content to  focus  on.  
%For example, if the value of relevance between sentence $i$ and sentence $j$, $\boldsymbol{h}_i^\mathrm{T}\boldsymbol{W}_r\boldsymbol{d}$, is less than the threshold $\epsilon_r$, the connection between them is masked, alternatively let $q_{ij} = 0$. 
Once satisfied with the sentence selections, the document graph is then reshaped   to align with the different mask matrices, and the summary selection is changed because of the  revised centrality. Generating the d ifferent summaries for the final output based on the controllable masks is done without additional training.  In turn, enforcing sentence-deployed attention through Eq.(\ref{eq:focus}) tells the abstractor what to focus on as it infers and generates the abstractive summary.

\section{Experiments}
\label{sec:exp}
 In this section, we describe the datasets used in the experiments, our setup, implementation details\footnote{Our code and dataset samples are available on \url{https://github.com/Wanghn95/Esca_Code}} and evaluation methods, and analyze the result.
 
\subsubsection{Datasets}
\label{sect:data}
We evaluated our models and baselines on two benchmark datasets, namely the CNN/DailyMail news set  \citep{hermann2015teaching}, and the New York Annotated Corpus (NYT) \citep{sandhaus2008new}. %, and the XSum \citep{narayan2018don}. 
 The {CNN/DailyMail} dataset\footnote{\url{https://cs.nyu.edu/~kcho/DMQA/}} contains news articles and associated highlights as summaries. We followed the standard splits  90,266/1220/1093 for the training, validation and testing sets for the CNN dataset and 196,961/12,148/10,397 for the DailyMail dataset. We did not anonymize the entities, and the datasets were pre-processed following \citet{DBLP:conf/acl/SeeLM17}. %Input documents were truncated to 500 tokens. 
 The {NYT} dataset\footnote{\url{https://catalog.ldc.upenn.edu/LDC2008T19}} contains  110,540 articles with abstractive summaries, which were divided into 100,834 articles for the training set and 9,706 for the test set, following \citet{durrett2016learning}.  We also filtered the raw datasets by eliminating the documents with summaries shorter than 50 words. The filtered test set, called  NTY50, included 3,421 examples. The abstractor processed the input by truncating the source documents to 400 tokens for CNN/DailyMail and 800 for NYT. 
As discussed in \citet{liu2019text}, the NYT test set contains longer and more elaborate summaries than the CNN/DailyMail set,  whose summaries are largely extractive and mostly concentrate on the beginning of the documents. All sentences were split with the Stanford CoreNLP toolkit \citep{manning2014}. 

We used ROUGE  as the evaluation metric \cite{lin2004rouge}\footnote{Implemented by pyrouge package based on ROUGE1.5.5.}, which measures the quality of a summary by computing the overlapping lexical elements between the candidate summary and a reference summary. Following  previous practice, we assessed R-1 (unigram), R-2 (bigram) and R-L (longest common subsequence). 

%\subsection{Details of Our Models}
\subsubsection{ESCA-Transformer} 
was trained with a 6-layer transformer. The hidden size was set to 512, and the feed-forward dimension for the multi-head attention was set to 1024. 8 heads. We used dropout with a probability of 0.2 prior to the linear layers. The learning rate for the pointer-generator was 0.15 with a batch size for the encoder of 32 and a beam size for the decoder of 4.  The learning rate of both the extractor and abstractor was 0.15. At the testing phase, we limited the length of the summary to 120 words. The model was trained with an early stopping and length penalty imposed on the validation set. 
%For the end-to-end training, the batch size and learning rate is same to the pre-training model, 48 and 0.15. 
%We used the beam search algorithm to generate multiple summary candidates in parallel to obtain better results.  
%To encourage the generation of longer sequences, we applied length penalty that is formulated as$\frac{(5+|y|)^\eta}{(5+1)^\eta}$, the  parameter $\eta$  is 0.9. 
%The minimum length of the generated summary is set to 35.

\subsubsection{ESCA-BERT}
  followed the settings specified by  \citet{liu2019text}. Specifically, we inserted \texttt{[CLS]} tokens at the start of each sentence, and also used two-interval segment embeddings \texttt{[$E_A$]} or  \texttt{[$E_B$]} to distinguish between multiple sentences in a document. The  \texttt{[CLS]} then learned the sentence embedding. Position embeddings in the BERT model had a 512 length limit. We  used the standard ‘BERT-base-uncased’ version of BERT \footnote{\url{https://github.com/google-research/bert}}.  Both the source and target tokens were tokenized with BERT’s subwords. The hidden size of the transformer layers was 768, and all the feed-forward layers had 2048 hidden units. One transformer layer in the extractor with 8 heads and a dropout of 0.1 was dedicated to producing the sentence representations. We used the trigram block trick \citep{paulus2017deep} to prevent duplicates. The abstractor was trained over 15k iterations for the NYT dataset and 100k iterations for CNN/DM with label smoothing loss  \citep{szegedy2016rethinking} at a factor of 0.1. Moreover, dropout with a probability of 0.2 was applied prior to the linear layers. 
The decoder contained 6 transformer layers. We used separate learning rates of 0.002 and 0.2 for the BERT encoder and Transformer decoder, respectively. The settings for the decoding process were the same as those outlined for the Transformer-based model above. 
%We chose the best  checkpoint for the testing through a validation process, and conducted the experiments on 2 pieces of 2080Ti GPUs. Training took 2 days for the abstractor, 1 day for the extractor, and 1 day for the hybrid ESCA model.
The final model contained 180M parameters. %We used cross-validation to choose hyper-parameters.

\subsubsection{Comparative Models}
Each of the following state-of-the-art models follow the ``select and  generate" style (BERTSUMabs excluded), as Section outlined in related work. The eight chosen comparators were:  \textsc{PG+Coverage} is a BiGRU-based seq2seq model integrated with a pointer network and  an additional coverage  mechanism \citep{DBLP:conf/acl/SeeLM17}. 
%\textsc{Graph-attention} \citep{tan2017abstractive} employed a graph ranking-based attention to identify salient sentences. 
\textsc{Select-Reinforce} \citep{chen2018fast}, which reinforces the extraction of important sentences with a reward function based on a summary rewrite evaluation metric.  
\textsc{Inconsistency-Loss} \citep{hsu2018unified} includes a loss function that uses sentence-level attention to modulate word-level attention for generating summaries. 
\textsc{Bottom-Up} \citep{gehrmann2018bottom}
uses an extractive encoder as a content selector to constrain word attention for the abstractive summarization. 
\textsc{ExplicitSelection}  \citep{li2018improving} is an extended version of the vanilla seq2seq model with a soft information selection layer to control information flow. 
\textsc{SENECA} \citep{sharma2019entity} selects entity-aware sentences and then connects them abstract generation based on reinforcement learning. 
 %\textbf{ETADS} \citep{you2019improving} uses a focus-attention layer and a salience-selection layer with the transformer-based encoder-decoder framework.  
\textsc{BERTSUMabs} and \textsc{BERTSUMextabs} are developed by  \citet{liu2019text}, which are not ``select-and-generate models". \textsc{BERTSUMextabs} adopts a two-stage fine-tune of extractor and abstractor. 
 %In addition, \textbf{Oracle} in this paper meant the abstract summary that was  generated from the labeled ground-truth sentences. 
  
 %Self-designed baselines include models that adopt the same framework of the proposed select and generate strategy but with different forms of encoders and decoders. % \textbf{BiGRU+Point} is GRU-based encoder-decoder framework and the extractor employs  point-wise ranking strategy, while \textbf{BiGRU+Pair} employs pair-wise extractor. 
 
 %In this paper, we propose the ESCA, including  ESCA-Transformer and ESCA-BERT, respectively. The details of experimental model settings are described in Appendix D. 

\subsection{Quantitative Analysis}
The overall results are presented in Table \ref{tab:cnn/dm} and \ref{tab:nyt}. We observed that \textsc{ESCA-Transformer} had competitive performance to most of the baselines, and \textsc{ESCA-BERT} outperformed all the strong state-of-the-arts on both datasets in all metrics. Relatively speaking,  \textsc{ESCA-BERT} has a higher improvement (1.20\% comparing with the most advanced BERTExtAbs) in R-2 metric on the NYT dataset whose gold summaries are longer and more abstractive than the ones in the CNN/DailyMail datasets \citep{liu2019text}. It indicates that the ESCA model has advantage on generating long-length and fluent summaries. 
%It further enunciates the transformer structure may not be suitable for sequence generation unless the representation is augmented by the pre-trained language model. 

\begin{table}[t]
   \centering
   \resizebox{\linewidth}{!}{
    \begin{tabular}{l|c c c}
    \hline
   \textbf{Models} & \textbf{R-1} & \textbf{R-2} & \textbf{R-L}\\ \hline
    % LEAD-3 & 40.42 & 17.62 & 36.67\\
    % \textsc{Oracle} & 52.59 & 31.23 & 48.87\\
     %\hline
 %   \textsc{PG} & 36.44 & 15.66 & 33.42 \\ 
    \textsc{PG+Coverage} & 39.53 $^\ast$& 17.28$^\ast$ & 36.38$^\ast$ \\
  %  \textsc{Graph-attention} $\dag$ & 38.1  &13.9  &34.0\\
    \textsc{Select-Reinforce}   & 40.88 $^\ast$&17.80$^\ast$ &38.54$^\ast$\\
    \textsc{Inconsistency-Loss}   & 40.68$^\ast$ &17.97$^\ast$ &37.13$^\ast$\\
    \textsc{Bottom-Up}   & 41.22$^\ast$ & 18.68$^\ast$ & 38.34$^\ast$\\
    \textsc{Explicit-Select} $\dag$ & 41.54 & 18.18  & 36.47 \\
    \textsc{SENECA} $\dag$ & 41.52 & 18.36 & 38.09\\
   % ETADS $\dag$ & 41.75 &19.01 &38.89\\
  %  \hline
    \textsc{BERTSUMabs}  & 41.72$^\ast$ & 19.39$^\ast$ & {38.76}$^\ast$\\
  %  BERTSUMextabs $\dag$ &  42.13 & 19.60 & 39.18\\
    \hline
    \textbf{Ours}\\
    \textsc{ESCA-Transformer} &  41.34$^\ast$ & 18.50$^\ast$ & 37.94$^\ast$ \\
    \textsc{ESCA-BERT} & \textbf{42.01} & \textbf{19.52} & \textbf{39.07} \\
    %  修改前Transformer $\dag$ &  41.65 & 18.89 & 37.94\\
    %  修改前BERT $\dag$ &  42.12 & 19.52 & 39.07\\
    % \hline
    %  Oracle   & 45.84 & 22.56 & 42.67\\
   \hline     
    \end{tabular}
    }
\caption{ The ROUGE scores for the abstractive summaries from the CNN/DailyMail datasets. Results marked with a $\dag$ mark are taken from the their corresponding papers. $\ast$  indicates a significant difference between the comparing model and the ESCA-BERT (at $p<0.1$, using a pairwise t-test).}
    \label{tab:cnn/dm}
\end{table}

\begin{table}[t]
   \centering
   \resizebox{\linewidth}{!}{
    \begin{tabular}{l|c c c}
    \hline
   \textbf{Models} & \textbf{R-1} & \textbf{R-2} & \textbf{R-L}\\ \hline  
    % LEAD-3 & 40.42 & 17.62 & 36.67\\
    % \hline
  %  \textsc{PG} & 42.47& 25.61 & 36.10 \\ 
    \textsc{PG+Coverage} & 43.71$^\ast$ &26.40 $^\ast$& 37.79$^\ast$ \\
    \textsc{Bottom-Up}  & 47.38 $^\ast$&31.23$^\ast$& 41.81$^\ast$\\
    \textsc{SENECA} $\dag$ & 47.94 &31.77& 44.34\\
   % ETADS & 47.75 &19.01 &38.89\\
   % \hline
    \textsc{BERTSUMabs}   & 48.92$^\ast$ &30.84$^\ast$ &45.41$^\ast$\\
    \textsc{BERTSUMextabs}  &  49.02$^\ast$ &31.02$^\ast$& 45.55$^\ast$\\
    \hline
    \textbf{Ours}\\
    \textsc{ESCA-Transformer} &  47.63$^\ast$ & 30.10$^\ast$& 43.94$^\ast$\\
    \textsc{ESCA-BERT} &  \textbf{49.41} & \textbf{32.22} & \textbf{45.83}  \\
    \hline
   %  指标没有变化
   %  Oracle   & 53.93 & 35.64 & 49.18\\
   %\hline     
    \end{tabular}
    }
\caption{The ROUGE scores for the  abstractive summaries from the NYT50. Results marked with a $\dag$ are taken from their corresponding papers. $\ast$  indicates a significant difference between the comparing model and 
the ESCA-BERT ($p<0.1$, using a pairwise t-test).
}
    \label{tab:nyt}
\end{table}

\subsubsection{Ablation Studies of the Extractor}
\begin{table}[t]
   \centering
   %\resizebox{\linewidth}{!}{
    \begin{tabular}{l|c c c}
    \hline
   \textbf{Models} & \textbf{R-1} & \textbf{R-2} & \textbf{R-L}\\ \hline
    Extractor$_\text{PointWise}$ & 32.68  & 15.41  &  30.33\\
    Extractor$_\text{PairWise}$ & 36.41  & 17.19   & 33.68   \\
   \hline     
       Extractor$_\text{self-attention}$ & 42.8   & 20.1  & 39.2  \\
    Extractor$_\text{interaction-matrix}$ &  42.7  & 20.0  &  39.2   \\
    \hline
    \end{tabular}
  %  }
\caption{ROUGE scores from CNN/DailyMail datasets for the extractive summaries of ESCA-BERT model and its counterparts . %Top 6 sentences were selected in the first block.  Top 3 sentences were selected in the second block.
}
    \label{tab:pair-wise}
\end{table}
To investigate the effectiveness of our  pair-wise ranking strategy, we compared it with its counterpart - point-wise ranking. The probability of a sentence as a extractive summary was  calculated by  $\sigma(c_i)$ where $c_i$ is derived from Eq. (\ref{equ:centrality}). The point-wise ranking loss was then computed through the cross entropy of the predicted score with the gold label. The upper block of Table \ref{tab:pair-wise} focuses on the   extraction performance of the CNN/DailyMail dataset. 
%Regarding the analysis of the proposed pair-wise ranking  extractor and the often-used point-wise ranking extractor as \citet{nallapati2017summarunner,hsu2018unified, liu2019text}, we solely compared the extraction performances in Table \ref{tab:pair-wise} on the CNN/DailyMail dataset. 
It is clear that our model with pair-wise ranking is largely superior to the point-wise extractor in terms of the ROUGE scores, with  relative improvements ranging from 11.0\% to 11.55\%. %The reason  we used the recall metric is it's consistent with the extraction ground-truth label strategy in this paper. 
Since our study focuses more on the global sentence distributions over the documents, we selected the top 6 sentences for a more granular analysis of the results. Our observations suggest that the overall distribution of the pair-wise extractor is more likely to be close to the gold summary. This verifies that the pair-wise ranking has a noticeable effect on the quality of the extracted summary.  

% \begin{table}[t]
%   \centering
%   %\resizebox{\linewidth}{!}{
%     \begin{tabular}{l|c c c}
%     \hline
%   \textbf{Models} & \textbf{R-1} & \textbf{R-2} & \textbf{R-L}\\ \hline
%     Extractor$_\text{self-attention}$ & 42.8   & 20.1  & 39.2  \\
%     Extractor$_\text{interaction-matrix}$ &  42.7  & 20.0  &  39.2   \\
%   \hline     
%     \end{tabular}
%   %  }
% \caption{ROUGE scores of ESCA-BERT model and its counterpart (self-attention) on the CNN/DailyMail dataset. Top 3 sentences were selected.  }
%     \label{tab:self-attent}
% \end{table}
According to \citet{vaswani2017attention},  inter-sentence relations can also be captured by multiple stacked  self-attention layers. Therefore, we replaced the interaction matrix with a 2-layer self-attention mechanism to build a counterpart variant of  ESCA's extractor, called Extractor$_\text{self-attention}$. We selected the top 3 sentences for evaluation and, report the comparative  results in the lower block of Table  \ref{tab:pair-wise}. Based on the scores, there was no significant difference in the quality of the selection. However,  self-attention may not be able to adequately explain why each sentence was selected from the interaction matrix.

%if our proposed matrix can further outperform the self-attention operation in terms of ROUGE score? From Table, we cannot find much difference 

\subsection{Controllability}

\subsubsection{Synthetic Datasets} 

\begin{table}[t]
   \centering
   \resizebox{\linewidth}{!}{
    \begin{tabular}{ll|c c c}
    \hline
   \textbf{Control} & \textbf{Threshold} &\textbf{R-1} & \textbf{R-2} & \textbf{R-L}\\ \hline
   Novelty & $\epsilon_n=0$   & 44.78    & 35.39   & 42.25  \\  
  &  $\epsilon_n=0.3$  &  45.66 $\uparrow$  &  36.28 $\uparrow$ &  43.05 $\uparrow$ \\  
   &  $\epsilon_n=0.4$  &  45.26 $\uparrow$ &  36.08 $\uparrow$ &  42.67 $\uparrow$ \\ 
    &  $\epsilon_n=0.5$  &  45.28 $\uparrow$  &  35.90 $\uparrow$ &  42.71 $\uparrow$ \\ \hline
    Relevance & $\epsilon_r=0$   & 41.35    & 18.50   & 38.57  \\  
  &  $\epsilon_r=0.3$  &  41.41  $\uparrow$ &  18.57 $\uparrow$ &  38.62 $\uparrow$ \\  
   &  $\epsilon_r=0.5$  &  41.52  $\uparrow$ &  18.67 $\uparrow$ &  38.55 $\downarrow$ \\ 
    &  $\epsilon_r=0.7$  &  41.27 $\downarrow$  &  18.44 $\downarrow$ &  38.43 $\downarrow$ \\ 
    \hline
    \end{tabular}
    }
\caption{Controllability:  the ROUGE scores from the CNN/DailyMail datasets for different thresholds of  novelty $\epsilon_n$ and relevance $\epsilon_r$ (absolute decrease/increase performance over $\epsilon_n=\epsilon_r=0$ is denoted by $\uparrow$/$\downarrow$).  }
    \label{tab:control}
\end{table}
To evaluate the impact of attending to relevance and novelty, we created two sample datasets based on the testing set of the two original CNN/DailyMail sets. The dataset for relevance  was constructed by adding a  title as part of the original gold summaries to increase the relevance between the summaries and the input document. In terms of novelty \citet{zhou2018neural} found that the CNN/DailyMail gold summaries favor leading sentences, which may not cover the   content of the document comprehensively. Hence, we employed an advanced unsupervised extractive summarization method called,  \textsc{PacSum}   \cite{zheng2019sentence}, to discover more diverse summaries. \textsc{PacSum} disregards the first five sentences in an article and then selects the top-3-ranked sentences from the remainder of the input document for extraction. Then, the original gold summaries are complemented with the novel content. 
To explore the explainable selection w.r.t relevance and novelty as mentioned in Eq.(\ref{equ:interQ}), we manually set different thresholds to construct the masking matrices with  Eq.(\ref{equ:control}).   We used ROUGE F1 to evaluate the influence of these controllable thresholds on the two synthetic datasets. The scores are shown in Table  \ref{tab:control}. 
The results illustrate the control over different scales of novelty is indeed able to generate diverse summaries, while a relevance score of $\epsilon_r=0.5$ (except for R-L) generated the best summaries. However,  there is always a trade-off between controllability and summary quality.

It also shows that the ROUGE scores are varied weakly because of two reasons. First, the controlled summaries must  preserve the informative content that the original ESCA has. The ROUGE score cannot be largely changed, otherwise, the summary can be wrong. Second, ROUGE score has drawbacks for evaluation w.r.t the overlapped vocabularies.   
To further verify the effectiveness of these two controllable parameters, we conducted a human evaluation with novelty and relevance criteria in the following section and  some examples are provided in  Appendix C for further inspection.
  
\subsubsection{Explainable Matrix over Control}

% \begin{figure}[t]
%     \centering
%     \subfigure[]{\includegraphics[width =4cm]{1.png}\label{a1}}
%     \subfigure[]
%     {\includegraphics[width =3.8cm]{2.png}\label{a2}}
%     \subfigure[]
%     {\includegraphics[width =3.8cm]{3.png}\label{a3}}
%     \caption{Visualization of the interaction matrix $Q_s$ (Figure \ref{a1}), and reshaped matrix controlled by novelty (Figure \ref{a2}) and relevance (Figure \ref{a3}) metrics, respectively. We set the thresholds $\epsilon_n = \epsilon_r = 0.5$ for simplicity. }
%     \label{fig:visualQ}
% \end{figure}

\begin{figure}[t]
	\centering
     \includegraphics[width=0.45\textwidth]{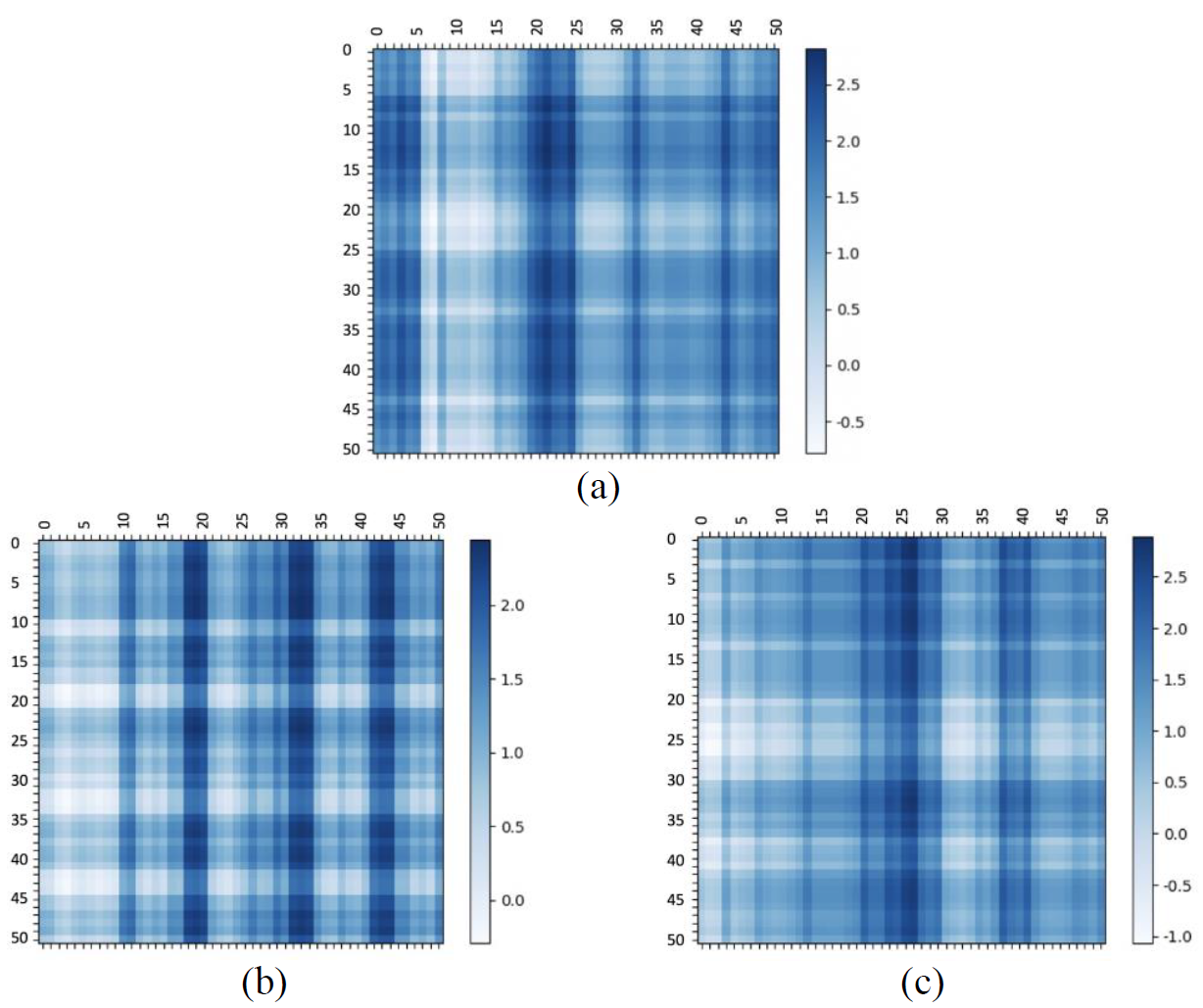}
	\caption{(a) A visualization of the interaction matrix $Q_s$. (b) The  reshaped matrix controlled by novelty. (c) the reshaped matrix according to relevance. For simplicity, the thresholds were set to $\epsilon_n = \epsilon_r = 0.5$.}  
	\label{fig:visualQ}
\end{figure}
\vspace{-4pt}
To explicitly demonstrate the power of the interaction matrix $\mathbf{Q}_s$ how the influence of novelty and relevance explains the final abstractive summaries, we visualize them as heatmaps, shown in Figure  \ref{fig:visualQ}. From these, we find that novelty can move the focused centrality from leading sentences to scattered spans of the document which capture  novel content (Figure \ref{fig:visualQ}(b)). Relevance slightly decreases the effect of leading sentences, while enhancing the centrality of salient content (Figure \ref{fig:visualQ}(c)).  

 \vspace{-4pt}
\subsection{Human Evaluation}
\label{sec:human}
Two separate human evaluations were conducted. The first was a question answering (QA) test and the second was to assess the quality of the summaries. 
Both types of evaluations were conducted on the Amazon Mechanical Turk platform. %We elicited 3 responses per HIT.  

\textbf{QA evaluation}  have been previously used to evaluate the quality of a summary and document compression \citep{clarke-lapata-2010-discourse,narayan2018ranking}.  It quantifies degree to which summarization models retain key information from the document. The more questions a system can answer, the better it is at summarizing the document.  Following a similar paradigm, we devised a set of questions (normally 2-3 question per summary) based on the gold summary on the assumption that those summaries did, in fact, highlight the most important content in the document content.  Appendix B shows some example QAs. Participants  were asked to answer the given questions by only reading the summaries without access to the source articles.  We elicited 3 responses per HIT.   
Similar with \citet{clarke-lapata-2010-discourse}, we  let the evaluation score be marked with 1 if the answer was correct, 0.5  if   partially correct, and 0 otherwise. %Each summary  elicited  3 responses.

%\vspace{-4pt}
\textbf{Criteria ranking}
To assess the quality of the summaries, we gave the participants the full article and asked them to select the best and worst summaries from the original summary in the dataset and those produced by each model according to four specific criteria:  
\textit{Informativeness} (how much useful information does the summary provide?), \textit{Novelty} (how much new information does each summary sentence provide?), \textit{Relevance} (how well the summary is relevant to input document?), and  \textit{Fluency} (how well the summary sentences are grammatically correct or easy to read?). We randomly select 20 instances from CNN/DM dataset to conduct the criteria ranking. 
The scores were computed as the percentage of times a summary was chosen as the best minus the times it was selected as the worst. The scores range from -1 (worst) to 1 (best).
\begin{table}[t]
	\centering
	\resizebox{\linewidth}{!}{
	\begin{tabular}{l|c|cccc}
		\hline
		Models &  QA         & \multicolumn{4}{ c }{Criteria} \\ 
		\cline{3-6}&& Infor. & Nov.& Rel.& Flu.\\
		\hline
		\textsc{PG+Cov.} & 26.0$^\ast$& -0.28 $^\ast$& -0.43 $^\ast$& -0.05 $^\ast$& -0.39$^\ast$\\ 
		\textsc{Bottom-Up} & 31.3$^\ast$& -0.07$^\ast$ & 0.02 $^\ast$& -0.08 $^\ast$& -0.02$^\ast$\\ 
		\textsc{Inconsistency} & 29.8$^\ast$& -0.10$^\ast$ & -0.12$^\ast$& -0.15$^\ast$ & -0.14$^\ast$\\
        \textsc{ESCA-BERT} & 39.2& 0.15 & 0.14 & 0.15 & 0.12 \\
        \textsc{Gold} & $\flat$& 0.30 & 0.40 & 0.13 & 0.48\\
    \hline
    \textsc{Bottom-Up} & $\flat$ & -0.23 &  -0.07  & -0.15 & $\flat$\\
     \textsc{ESCA-BERT} & $\flat$&  \textbf{0.10}  & 0.03  & 0.05  & $\flat$\\
     ESCA($\epsilon_n=0.3$) & $\flat$&  0.05  &  \textbf{0.10} & 0.02 &$\flat$\\
     ESCA($\epsilon_r=0.5$) & $\flat$&  0.07  & -0.02  &  \textbf{0.07} &$\flat$\\
          
          \hline
	\end{tabular}
}
	\caption{ QA and criteria-based human  evaluation. $\ast$  indicates statistically significant improvements over the baselines with ESCA-BERT (from a paired t-test at $p<0.05$).  Gold summaries were not included in QA evaluation. $\flat$ means it does not need to be evaluated by the specific use.
	} 
	\label{tab:humaneva} 
\end{table}

Based on the QA evaluation in Table \ref{tab:humaneva}, the summaries produced by ESCA-BERT spanned significantly more salient content. In the first block of the criteria ranking, the gold summary sets the upper bound, except for relevance.  Unsurprisingly, since the gold summaries of CNN/DailyMail are mostly from the top sentences in the articles, their relevance cannot be guaranteed. We also found that ESCA-BERT produced the most  popular summaries comparing with the other baseline in terms of the four criteria metrics. In the second block, ESCA with novelty and relevance controls were evaluated together with the \textsc{Bottom-up} and original ESCA. The difference in rankings varied slightly but, overall, the results clearly  prove that ESCA with controllable novelty or relevance gained the highest rank at corresponding criteria ($\epsilon_n$ or $\epsilon_r$, 
bold the highest value in the second block of Table \ref{tab:humaneva}). 
 \vspace{-4pt}
\section{Conclusion}
\label{sec:con}
This paper presents a novel hybrid framework for document summarization. The proposed ESCA is hybrid model equipped with a pair-wise ranking extractor that seamlessly connects with an abstractor armed with a sentence-level attention pointer. The flow of the framework is designed to explicitly explain why sentences are marked for extraction and to allow the operator to control exactly which sentences are ultimately extracted according to novelty and relevance scores. The subsequent abstractive generation process attends to these metrics when inferring the final summaries to produce the most desirable result.  Both empirical and subjective experiments show that our model makes a statistically significant improvement over stat-of-the-art baselines.

%This paper presents a novel hybrid framework long document summarization over long-range contexts. The proposed ESCA can explicitly explain the selection and control of the abstract generation via the metrics of relevance and novelty. It is equipped with the  proposed pair-wise ranking extractor and seamlessly connects with the abstractor with the sentence deployed pointer. Therefore, the generation can be affected through the updated deploying attention. Both empirical and subjective experiments show that our model makes a statistically significant improvement over state-of-the-art baselines.

 \section*{Acknowledgments}
This work was supported by the National Key Research and Development Program of China (Grant No. 2016YFB1000902) and the National Natural Science Foundation of China (Grant No. 61751201, U19B2020).

\appendix
\section{Appendix A: Extractor Labels}
\label{app:label}
The extractor was trained through a sentence-level binary classification   for each document.  We employed a greedy search to select the best combination of sentences that maximized ROUGE-L Recall and ROUGE-L F1 with reference to the human summary since our model required more salable and novel content for generation. The greedy search approach have been used in \citep{sharma2019entity, nallapati2017summarunner, hsu2018unified}, but each of them utilised  different ROUGE metrics, such as ROUGE-2 or ROUGE-1.  
 We stopped when none of the remaining candidate
sentences improves the ROUGE score upon addition to the current summary set. We returned this subset of sentences as the extractive ground-truth.

\section{Appendix B: Example  of QA Human Evaluation}
\label{sec:humanevaluation}

In our experiment, the QA-based human evaluation is   created to testify  that if  the generated summary is able to answer the questions that are drawn from its gold summary. In this way, the relevance, readability and informativeness can be evaluated. In Table  \ref{fig:examplehuman}, we show an example of a sample gold summary  and its questions with answers. 

When taken the QA-evaluation, the summaries that were generated by different systems and the corresponding questions  were presented to the participants at the Amazon’s Mechanical Turk crowdsourcing platform.  

\begin{table}[!htb]
\centering
\begin{tabular}{|p{0.45\textwidth}|}
\hline CNN/DailyMail\\ \hline
\textbf{Gold Summary}: Emergency services were called to the Kosciuszko Bridge at about 11.50 am Monday , where a woman had climbed over the bridge 's railing and was standing on a section of metal piping. Officers tried to calm her down as NYPD patrol boats cruised under the bridge on Newtown creek , which connects Greenpoint in Brooklyn and Maspeth in queens. A witness said the woman was a 44-year-old polish mother-of-one who was going through a tough divorce. She agreed to be rescued after police talked to her about her daughter and was taken to elmhurst hospital.\\
 \hline
\textbf{Questions}: \\
$\bullet$  When was Emergency services called to the Kosciuszko Bridge? \\Answer: 11.50 am\\
$\bullet$  What did the witness say about the women?\\ Answer: 44-year-old Polish mother-of-one who was going through a tough divorce\\
$\bullet$ Did the women agreed to be rescued?\\ Answer: Yes\\
\hline 
\end{tabular}
\caption{\label{fig:examplehuman} One Example gold summary from our human evaluation set and questions with answer key created for it. The examples are selected from  CNN/Daily Mail.}
\end{table}

\section{Appendix C: Examples of Controllability}
To  demonstrate  the  effectiveness of the controllability proposed in this model, this  appendix  provides  examples with respect to relevance control $\epsilon_r$ and novelty control $\epsilon_n$ in Table \ref{example},  with  side-by-side  comparisons of  gold summary and  the  summaries produced  by the ESCA model and corresponding controlled model. 

As the NO.1 Relevance shows, the  controlled model is able to generate more relevant content with respect to the gold summary, highlighted by green color. The relevance control generates more relevant summary, in the meanwhile,  preserve  the content that the original ESCA has generated (underlined in all the included  sentences).

%Since automatic evaluation produces bias based on the annotated gold summary, the ROUGE metrics of the controlled models cannot reveal the benefits of the controllability. 

%
\begin{table*}[!htb]
\centering
\begin{tabular}{|p{0.9\textwidth}|}
\hline 
\textbf{NO.1: Relevance ($\epsilon_r=0.5$)}\\
\hline
\textbf{Gold Summary}\\
$\bullet$  \uuline{Police officers have shut down an enormous 1000 rave in Sydney's east.}\\
$\bullet$ They were called to abandoned industrial area in botany on Saturday night.\\
$\bullet$ \colorbox{green}{Police were forced to use capsicum spray on the group} after back up came.\\
$\bullet$ \uuline{One officer had glass removed from his headafter the crowd threw bottles} .\\
$\bullet$ \colorbox{green}{A woman was arrested and is being questioned after assaulting an officer}.
\\\hline
\textbf{ESCA}\\
$\bullet$ Two police officers have sustained injuries after attempting to close \uuline{down an enormous 1000 rave in Sydney's east}.\\
$\bullet$ \uuline{One officer had to have a piece of glass removed from his head after having a bottle thrown at him}.\\
$\bullet$ Police were assisted by back up officers as well as the riot squad and dog squads.\\
\hline
\textbf{Control by Relevance}\\
$\bullet$ Two police officers have sustained injuries after attempting to close \uuline{down an enormous 1000 person rave in Sydney's east}.\\
$\bullet$ At about 10.30 pm on Saturday night, police received a number of complaints about a dangerously large party at an abandoned industrial area on McPherson street in botany.\\
$\bullet$ \colorbox{green}{Police were forced to use capsicum spray on} a number of the attendees and \uuline{one officer had to have a piece of glass removed from his head after having a bottle thrown at him}.\\
$\bullet$ \colorbox{green}{A 26-year-old woman was arrested after she allegedly assaulted an officer}.
\\
\hline
\hline
\textbf{NO.2: Novelty}($\epsilon_n=0.3$)\\
\hline
\textbf{Gold Summary}\\
$\bullet$ \uuline{Jeralean Talley was born on may 23, 1899}.\\
$\bullet$ \uuline{She credits her longevity to her faith}.\\
$\bullet$ \uuline{Inherited the title of world's oldest person following the death of Arkansas woman Gertrude Weaver, 116, }on Monday.\\
\hline
\textbf{ESCA}\\
$\bullet$ \uuline{Jeralean Talley was born in rural montrose on may 23, 1899 , and credits her long life to her faith}.\\ 
$\bullet$ \uuline{Asked for her key to longevity}, the Detroit free press reports that she echoed previous answers on the topic.\\
$\bullet$ Gertrude Weaver, a \uuline{116-year-old arkansas woman who was the oldest documented person} for a total of six days.\\
\hline
\textbf{Control by Novelty}\\
$\bullet$ Jeralean Talley of Inkster tops a list maintained by the Los Angeles-based Geron planck research group , \uuline{which tracks the world's longest-living people}.\\
$\bullet$ \uuline{Talley was born on may 23, 1899 , and credits her long life to her faith.}\\
$\bullet$ \colorbox{pink}{Talley's five generations of her family have lived in the Detroit area}.\\\hline
\end{tabular}
\caption{\label{fig:examplecontrol} Examples of summaries that have been controlled under relevance and novelty criteria. The summaries include gold summary, the summary from the ESCA and the corresponding summary upon specific controls}
\label{example}
\end{table*}

The NO.2 Novelty deals with a different problem. According to our experience on ROUGE metrics and human evaluation, we found that novelty is the criteria that the current automatic metric cannot well evaluate. Take the example in NO.2 Novelty, both gold summary and the summary generated by the original ESCA model provide very similar topics --- ``Talley's longevity" (We underlined the similar topics in the table). 
But a good quality of summary should cover new topics of the content. 
 Surprisingly, the model under novelty control generated a brand new sentence that is from  the source document but is not ``Talley's longevity" any more, which we highlighted it with red color. 
This is just the reason why our model under control gained the better performance in terms of human evaluated novelty criteria. 

\bibliography{aaai21_arxiv}

\end{document}